\documentclass[11pt,a4paper]{article}
\usepackage[hyperref]{emnlp2020}
\usepackage{times}
\usepackage{latexsym}
\usepackage{amsmath}
\usepackage{amssymb}
\usepackage{float}
\usepackage{booktabs}
\usepackage{enumerate}
\usepackage{tikz} 
\usetikzlibrary{calc}

\DeclareMathOperator{\softmax}{softmax}

\usepackage{tikz}
\usepackage{pgfplots}
\usetikzlibrary{shapes, arrows, positioning, decorations.markings, shadows,shapes.geometric}

\definecolor {processblue}{cmyk}{0.96,0,0,0}
\tikzstyle{int}=[draw, fill=blue!20, minimum size=2em]
\tikzstyle{init} = [pin edge={to-,thin,black}]

\usetikzlibrary{shapes}
\usetikzlibrary{fit}
\usetikzlibrary{chains}
\usetikzlibrary{arrows}

\tikzstyle{plate} = [draw, rectangle, rounded corners, fit=#1]
\tikzstyle{wrap} = [inner sep=0pt, fit=#1]

\tikzstyle{caption} = [node distance=0] %
\tikzstyle{bottom plate caption} = [caption, node distance=0, inner sep=0pt,
below left=-5pt and 0pt of #1.south east] %

\tikzstyle{top plate caption} = [caption, node distance=0, inner sep=0pt,
below left=0pt and 0pt of #1.north east] %

\usepackage{microtype}

\aclfinalcopy 

\title{Parallel Interactive Networks for Multi-Domain Dialogue State Generation}
\author{
  Junfan Chen \\
  BDBC and SKLSDE\\
  Beihang University, China \\
  \texttt{chenjf@act.buaa.edu.cn} \\
  \And
  Richong Zhang\thanks{$\ \ $Corresponding author}\\
  BDBC and SKLSDE\\
  Beihang University, China \\
  \texttt{zhangrc@act.buaa.edu.cn} \\
  \AND
  Yongyi Mao \\
  School of EECS\\
  University of Ottawa, Canada \\
  \texttt{ymao@uottawa.ca} \\
  \And 
  Jie Xu \\
  School of Computing\\
  University of Leeds, United Kingdom \\
  \texttt{j.xu@leeds.ac.uk} \\
}  

\date{}
\begin{document}
\maketitle
\begin{abstract}
The dependencies between system and user utterances in the same turn and across different turns are not fully considered in existing multi-domain dialogue state tracking (MDST) models. In this study, we argue that the incorporation of these dependencies is crucial for the design of MDST and propose Parallel Interactive Networks (PIN) to model these dependencies. Specifically, we integrate an interactive encoder to jointly model the in-turn dependencies and cross-turn dependencies. The slot-level context is introduced to extract more expressive features for different slots. And a distributed copy mechanism is utilized to selectively copy words from historical system utterances or historical user utterances. Empirical studies demonstrated the superiority of the proposed PIN model.

\end{abstract}

\section{Introduction}
Spoken dialogue system (SDS) is an application that can help users complete their goals efficiently. 
An SDS usually has a logic engine, called dialogue manager, which involves two main sub-tasks for determining how the system will respond to the users: dialogue state tracking and dialogue policy learning. The task we discuss in this paper is dialogue state tracking, which allows the system maintaining an internal representation of the state of the dialogue as the dialogue progress~\cite{Young:10}. 

Dialogue state tracking involving a single domain has been extensively studied and achieved much progress. As a more challenging task, Multi-domain dialogue state tracking (MDST) has been introduced in~\cite{Ramadan:18} and attracts much attention in the research community. 
Instead of only predicting the $(slot, value)$ pair, in MDST, a model is expected to predict the $(domain, slot, value)$ triplets for each slot in each domain. This task is a great challenge not only because of the large ontology involving $30$ slots and exceeding $4500$ values~\cite{Wu:19}, but also the mixed-domain nature of the dialogues and some complex cases involving cross-turn inference. 

\begin{figure}[ht]
	\begin{center}
		\begin{tabular}{c}
			\begin{tikzpicture}[-latex ,auto ,node distance = 1 cm and 1 cm, on grid ,
    	semithick ,				
    	bigbox/.style ={rectangle, rounded corners, minimum width=7cm, minimum height=5.2cm, text centered, text width=4.5cm, draw=black, fill=black!10},
    	turn/.style ={rectangle, rounded corners, minimum width=4cm, minimum height=0.9cm, text centered, text width=4cm, draw=black, fill=white, drop shadow}, 
    	sys/.style ={rectangle, rounded corners, minimum width=1cm, minimum height=0.9cm, text centered, text width=1cm, draw=black, fill=white, drop shadow},  
    	usr/.style ={rectangle, rounded corners, minimum width=1cm, minimum height=0.9cm, text centered, text width=1cm, draw=black, fill=white, drop shadow},      	
        leftblank/.style ={rectangle, rounded corners, minimum width=0.6cm, minimum height=0.9cm, text centered, text width=1cm, draw=white, fill=white},   
        topblank/.style ={rectangle, rounded corners, minimum width=0.6cm, minimum height=0.5cm, text centered, text width=1.5cm, draw=white, fill=white},       
        textcell/.style ={rectangle, minimum width=8cm, minimum height=0.5cm, text width=8cm, draw=white, fill=white},             
    	state/.style ={rectangle, rounded corners, minimum width=6.8cm, minimum height=0.9cm, text centered, text width=6cm, draw=black, fill=white, drop shadow}    	
    	] 
		\node[textcell] (b0) at (0, 0) {\small \textbf{u1}: I'm finding a cheap european \textcolor{cyan}{restaurant}.};    
		\node[textcell] (ba) [above =of b0] {\small \textbf{s1}: Can I help you?};	 
		\node[textcell] (b1) [below =of b0] {\small \textbf{s2}: There is a \textcolor{cyan}{Curry Garden}, and \textcolor{cyan}{hotel}?};   
		\node[textcell] (b2) [below =of b1] {\small \textbf{u2}: I need a \textcolor{cyan}{hotel} \quad with \textcolor{cyan}{free-wifi}.};  			
		\node[textcell] (b3) [below =of b2] {\small \textbf{s3}: There is an \textcolor{cyan}{Ashley} hotel. Anything else?}; 
		\node[textcell] (b4) [below =of b3] {\small \textbf{u3}: A taxi from the \textcolor{cyan}{retaurant} to the \textcolor{cyan}{hotel}.};
		
		\draw[red,line width=1, -] (-4.1,-0.08) arc (135:225:0.6); 	
		\draw[blue,line width=1, -] (-4.1,-1.08) arc (135:225:0.6); 	
		\draw[red,line width=1, -] (-4.1,-2.08) arc (135:225:0.6); 
		\draw[blue,line width=1, -] (-4.1,-3.08) arc (135:225:0.6); 
		\draw[red,line width=1, -] (-4.1,-1.0) arc (120:240:1.75); 
		\draw[thick,red, -] (-1.1, -0.75) -- (1.2, -0.15); 
		\draw[thick,blue, -] (0.15, -1.85) -- (0.9, -1.15); 
		\draw[thick,red, -] (-1.5, -2.8) -- (-1.85, -2.15);
		\draw[thick,blue, -] (-1.5, -3.15) -- (1.3, -3.85);
		\draw[thick,red, -] (-0.50, -3.85) -- (-1.5, -1.15);
\end{tikzpicture}
		\end{tabular}
	\end{center}
	\caption{The dependencies between the system utterances and user utterances in a multi-domain dialogue. The red lines imply cross-turn dependencies and the blue lines imply in-turn dependencies.}
	\label{fig:intro}
\end{figure}

Several models have been proposed for MDST task and proven to be successful~\cite{Mrksic:15, Ramadan:18, Goel:19, Eric:19, Lee:19, Wu:19}. Among these models, TRADE~\cite{Wu:19} achieves the state-of-the-art on the MultiWOZ 2.0 dataset (one of the standard MDST datasets) by encoding the entire dialogue history using a bidirectional GRU and incorporating soft-gated copy mechanism to generate the values. Inspired by TRADE, we purpose to build a more accurate and robust state generator PIN. The motivations of proposing PIN is in two aspects. 

One aspect is considering the interactive nature of the dialogues. The interaction of the user and the system is often organized by a question-answering style. It is common in dialogue state tracking that a domain or slot being specified by one of the user or system, then the value being answered by the other. For example, in the dialogue in Figure~\ref{fig:intro}, the user specifies a {\em Restaurant} domain, and the system answers a restaurant name {\em Curry Garden}.  As is shown in Figure~\ref{fig:intro}, there are two types of dependencies, in-turn dependencies and cross-turn dependencies, both contribute to discovering slot-value pairs. It is worth noting that some hard cases involving inference actually rely on cross-turn dependencies (e.g., the dependency between utterance \textbf{s2} and \textbf{u3} in Figure~\ref{fig:intro}). Thus a correctly modeling of these dependencies can improve slot-value extraction and cross-turn inference. In this work, we build an Interactive Encoder which completely accords with the dependencies expressed in Figure~\ref{fig:intro} to jointly model the in-turn dependencies and cross-turn dependencies. 

The interactive nature of dialogues also implies that the value for a slot tends to be specified frequently either by a system or by a user. For example, the values for slots involving names, such as {\em Restaurant-name} and {\em Hotel-name} are likely to be provided by the system. And the values for the slots like {\em Hotel-stay} (the days to stay) and {\em Hotel-people} (the number of people booking for) are usually provided by the user. This observation inspires our designing of the distributed copy mechanism, which allows the state generator choosing to copy words from either the historical system utterances or the historical user utterances.

The other aspect is the slot overlapping problem in MDST. Unlike single-domain DST, slot overlapping is common in MDST, and these overlapping slots share similar values. For example, both the {\em Restaurant} and {\em Hotel} domain have a slot {\em price range} that shares the same values. Under this condition, a generator without considering slot-specific features may mistakenly extract the value of one slot as the value of some other slot. To overcome the slot overlapping problem, we introduce a slot-level context in the state generator.

In summary, we propose a generation-based MDST model which takes into consideration of the interactive nature of dialogues and slot overlapping problem in MDST. The contributions of this work are as follows.
\begin{itemize}
    \item We propose an interactive encoding method with two parallel double-layer recurrent networks which can jointly model the in-turn dependencies and cross-turn dependencies. 
    \item We introduce the slot-level context into the state generator to accurately generate the values for overlapping slots. 
    \item We present a distributed copy mechanism to selectively copy words from either the historical system utterances or the historical user utterances.
\end{itemize}

\section{Problem Statement}
In multi-domain dialogue state tracking, the {\em state} is usually expressed as a set of $(domain, slot, value)$ triplets. The {\em domain} refers to the topics of the dialogue, such as the {\em Restaurant} domain, which indicates that the dialogue involves restaurant booking. The {\em slot} is an aspect of the user's goals, such as {\em food}, {\em area} and {\em pricerange} in the restaurant-booking dialogues. And the {\em value} is the user's specific interests, such as {\em chinese} value for {\em food} slot that indicates the user is interested in Chinese food. The dialogue state is maintained so as to track the progress of the dialogue. At each turn, the system generates a system utterance in natural language, and the user responds to the system with some sentences, referred to as user utterance. The objective of multi-domain dialogue state tracking is to predict the value of each $(domain, slot)$ pair at each turn given the historical system utterances and user utterances. In this paper, the multi-domain dialogue state tracking is treated as a sequence generation task, where each word of a value is generated from a state generator. 

\section{Methodology}
In this section, we introduce the proposed PIN model. The model consists of four components: Interactive Encoder, Slot-level Context, Value Generator and Slot Gate. We next describe each component in detail.

\subsection{Interactive Encoder}
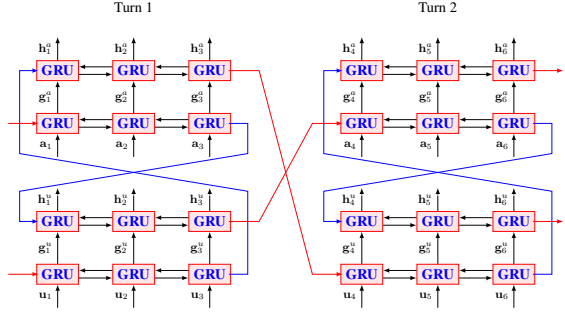
\begin{figure}[ht]
	\begin{center}
		\begin{tabular}{c}
			\scalebox{0.5}{
				\begin{tikzpicture}[-latex ,auto ,node distance =1.4 cm and 2 cm, on grid , semithick ,
    	state/.style ={ circle , draw=red, fill=red!10, text=red , minimum width =1cm},
    	blank/.style ={ circle , color =white ,
    		draw, white , text=black , minimum width =1cm},				
    	box/.style ={rectangle , draw=blue, fill=blue!30 ,
    		draw, text=blue , minimum width =0.5cm , minimum height = 0.5cm},
    	cell/.style = {rectangle , draw=red, fill=red!10 ,
    		 text=blue , minimum width = 1cm , minimum height = 0.5cm}]
red
    	\foreach \r/\c/\t/\s in {0/0/u/1,8/0/u/4,0/4/a/1,8/4/a/4}  
	    {
        	\coordinate  (P0) at (\r, \c);	
    		\path (P0) node[blank](b00){};
        	\node[blank](b01)[right=of b00]{};
            \node[blank](b02)[right=of b01]{};
        	\node[cell](c00)[below=of b00]{\bf GRU};
        	\node[cell](c01)[below=of b01]{\bf GRU};
        	\node[cell](c02)[below=of b02]{\bf GRU};
        	\node[cell](c10)[below=of c00]{\bf GRU};
        	\node[cell](c11)[below=of c01]{\bf GRU};
        	\node[cell](c12)[below=of c02]{\bf GRU};
        	\node[blank](b10)[below=of c10]{}; 
        	\path (b10) edge node[left]{\small$\mathbf{\t}_{\s}$} (c10);
        	\node[blank](b11)[below=of c11]{}; 
        	\pgfmathparse{int(\s+1)}
        	\path (b11) edge node[left]{\small$\mathbf{\t}_{\pgfmathresult}$} (c11);  
        	\node[blank](b12)[below=of c12]{}; 
        	\pgfmathparse{int(\s+2)}
        	\path (b12) edge node[left]{\small$\mathbf{\t}_{\pgfmathresult}$} (c12);        	
        	
            \path (c00) edge node[left]{\small$\mathbf{h}_{\s}^{\t}$} (b00);
            \pgfmathparse{int(\s+1)}
            \path (c01) edge node[left]{\small$\mathbf{h}_{\pgfmathresult}^{\t}$} (b01);
            \pgfmathparse{int(\s+2)}
            \path (c02) edge node[left]{\small$\mathbf{h}_{\pgfmathresult}^{\t}$} (b02);
            \path ([xshift=0cm,yshift=0.1cm]c01.west) edge  ([xshift=0cm, yshift=0.1cm]c00.east);
            \path ([xshift=0cm,yshift=-0.1cm]c00.east) edge  ([xshift=0cm, yshift=-0.1cm]c01.west);
            \path ([xshift=0cm,yshift=0.1cm]c02.west) edge  ([xshift=0cm, yshift=0.1cm]c01.east);
            \path ([xshift=0cm,yshift=-0.1cm]c01.east) edge  ([xshift=0cm, yshift=-0.1cm]c02.west);
            \path ([xshift=0cm,yshift=0.1cm]c11.west) edge  ([xshift=0cm, yshift=0.1cm]c10.east);
            \path ([xshift=0cm,yshift=-0.1cm]c10.east) edge  ([xshift=0cm, yshift=-0.1cm]c11.west);
            \path ([xshift=0cm,yshift=0.1cm]c12.west) edge  ([xshift=0cm, yshift=0.1cm]c11.east);
            \path ([xshift=0cm,yshift=-0.1cm]c11.east) edge  ([xshift=0cm, yshift=-0.1cm]c12.west);
            \path (c10) edge node[left]{\small$\mathbf{g}_{\s}^{\t}$} (c00);
            \pgfmathparse{int(\s+1)}
            \path (c11) edge node[left]{\small$\mathbf{g}_{\pgfmathresult}^{\t}$} (c01);
            \pgfmathparse{int(\s+2)}
            \path (c12) edge node[left]{\small$\mathbf{g}_{\pgfmathresult}^{\t}$} (c02);
        }
        \draw[blue] (4.5, 1.2) -- (5, 1.2) -- (5, 0.4) -- (-1, -0.6) -- (-1, -1.4) -- (-0.5, -1.4);
        \draw[blue] (4.5, -2.8) -- (5, -2.8) -- (5, -0.6) -- (-1, 0.4) -- (-1, 2.6) -- (-0.5, 2.6);
        \draw[red] (4.5, 2.6)--(5.3,2.6)--(6.7, -2.8)--(7.5,-2.8);
        \draw[red] (4.5, -1.4)--(5.3,-1.4)--(6.7, 1.2)--(7.5,1.2);
        
        \draw[blue] (12.5, -2.8) -- (13, -2.8) -- (13, -0.6) -- (7, 0.4) -- (7, 2.6) -- (7.5, 2.6);
        \draw[blue] (12.5, 1.2) -- (13, 1.2) -- (13, 0.4) -- (7, -0.6) -- (7, -1.4) -- (7.5, -1.4); 
        
        \draw[red] (-1.3, 1.2) -- (-0.5, 1.2);
        \draw[red] (-1.3, -2.8) -- (-0.5, -2.8);  
        \draw[red] (12.5, 2.6)--(13.3,2.6);
        \draw[red] (12.5, -1.4)--(13.3,-1.4);
        
        \path(2, 4) node[above]{Turn 1} ;
        \path(10, 4) node[above]{Turn 2} ;
    
\end{tikzpicture}
			}
		\end{tabular}
	\end{center}
	\vspace{-0.4cm}
	\caption{The structure of the Interactive Encoder. Due to space limitation, only two turns are shown. The red arrows emphasize modeling cross-turn dependencies and the blue arrows emphasize modeling in-turn dependencies.}
	\label{fig:interEncoder}
\end{figure}

Our design of the Interactive Encoder is inspired by the dependencies between the system and user utterances. Specifically, we wish to propose a novel network structure that completely represents the dependencies expressed in Figure~\ref{fig:intro}. A hierarchical recurrent networks with specific structures have been used to construct the Interactive Encoder, as shown in Figure~\ref{fig:interEncoder}.
The Interactive Encoder consists of two parallel hierarchical recurrent networks, one for historical system utterance encoding and another for historical user utterance encoding. The lower layer of the hierarchical recurrent networks allow each word to capture the cross-turn dependencies; and the higher layer of the hierarchical recurrent networks allows each word to capture the in-turn dependencies. In this way, the cross-turn dependencies and in-turn dependencies are jointly modeled.

We now present the details of the Interactive Encoder. Let $\mathbf{A}_l=\left\{ \mathbf{a}_{1}, \mathbf{a}_{2}, \cdots, \mathbf{a}_{m} \right\}$ denotes the sequence of word embeddings for the $l^{th}$ system utterance. And $\mathbf{U}_l=\left\{ \mathbf{u}_{1}, \mathbf{u}_{2}, \cdots, \mathbf{u}_{n} \right\}$ denotes the sequence of word embeddings for the the $l^{th}$ user utterance.  Here $m$ and $n$ denote the number of words in the $l^{th}$ system utterance and user utterance, respectively. 

For later use, we introduce a notation ${\rm GRE}(\mathbf{X},\mathbf{h}; \mathbf{W})$ to indicate the bi-directional GRU encoder~\cite{Chung:14} with inputs $\mathbf{X}$ (sequence of vector representations, such as word embeddings), parameters $\mathbf{W}$ and initialized hidden state $\mathbf{h}$.
The Interactive Encoder jointly models the cross-turn dependencies and in-turn dependencies through the following recurrent process.

The Interactive Encoder first let the input word embedding sequences $\mathbf{A}_l$ and $\mathbf{U}_l$ interact with the historical context, allowing the words capturing cross-turn dependencies
\begin{equation}
\begin{split}
\mathbf{G}_{l}^{a}, \mathbf{g}_{l}^{a} = {\rm GRE}(\mathbf{A}_l,\mathbf{h}_{l-1}^u; \mathbf{W}_a) \\
\mathbf{G}_{t}^{u}, \mathbf{g}_{l}^{u} = {\rm GRE}(\mathbf{U}_l,\mathbf{h}_{l-1}^a; \mathbf{W}_u)
\end{split}
\label{eq:enc1}
\end{equation}
where $\mathbf{W}_a$ and $\mathbf{W}_u$ are the parameters of the GRUs, the initialized hidden states $\mathbf{h}_{l-1}^a$ and $\mathbf{h}_{l-1}^u$ are respectively the system context vector and the user context vector generated from the last turn. $\mathbf{G}_{l}^{a}$ and $\mathbf{g}_{l}^{a}$ denote the entire sequence of output vectors and the last output vector of the GRUs, respectively.  

The outputs of the lower-layer GRUs, $\mathbf{G}_{l}^{a}$ and $\mathbf{g}_{l}^{a}$, are then feed into the higher-layer GRUs to interact with the current context for capturing in-turn dependencies 
\begin{equation}
	\begin{split}
		\mathbf{H}_{l}^{a
		}, \mathbf{h}_l^a = {\rm GRE}(\mathbf{G}_{l}^{a},\mathbf{g}_{l}^{u}; \mathbf{M}_a)\\		
		\mathbf{H}_{l}^{u},\mathbf{h}_l^u = {\rm GRE}(\mathbf{G}_{t}^{u},\mathbf{g}_{l}^{a}; \mathbf{M}_u)
	\end{split}
	\label{eq:enc2}
\end{equation}
where $\mathbf{M}_a$ and $\mathbf{M}_u$ are the parameters of the higher-layer GRUs, and $\mathbf{h}_{l}^a$ and $\mathbf{h}_{l}^u$ are the generated system context vector and user context vector of the current turn. $\mathbf{h}_{l}^a$ and $\mathbf{h}_{l}^u$ are then feed into the lower-layer GRUs as the initialized hidden states of the next turn. 

With this recurrent architecture, the Interactive Encoder captures the dependencies of the entire dialogue history by rolling from the first turn to the current turn. At the beginning of a dialogue, we set the initialized hidden states as zero vectors, that is $\mathbf{h}_0^a = \mathbf{h}_0^u = \mathbf{0}$.

The outputs from each turn of the dialogue are then concatenated as the system context sequence $\mathbf{H}^{a}=\left\{ \mathbf{h}_{1}^{a}, \mathbf{h}_{2}^{a}, \cdots, \mathbf{h}_{M}^{a} \right\}$ and user context sequence $\mathbf{H}^{u}=\left\{ \mathbf{h}_{1}^{u}, \mathbf{h}_{2}^{u}, \cdots, \mathbf{h}_{N}^{u} \right\}$. Here $M$ and $N$ denote the total number of words in historical system utterance and historical user utterance, respectively.
\begin{figure*}[ht]
	\begin{center}
		\begin{tabular}{c}
			\scalebox{0.7}{
				\begin{tikzpicture}
[-latex,auto,semithick,node distance =1 cm and 0.7 cm,
 blank/.style ={ circle , color =white ,
    		draw, white , text=black , minimum width =1cm},
 bar/.style = {rectangle , draw=black, fill=blue ,
 	text=black , minimum width = 0.2cm}, 
 vec/.style = {rectangle , draw=black, fill=yellow ,
  	text=black , minimum width = 0.3cm}, 
 vvec/.style = {rectangle , draw=black, fill=black!20 ,
 	text=black , minimum height = 0.05cm},   	
 cell/.style = {rectangle , draw=black, fill=white ,
    		 text=black , minimum width = 0.5cm , minimum height = 1cm}]
 \node[cell](a1) at (-0.1, 0) {};
 \draw (-2, 0) -- node[above]{U} (-0.7, 0);
 \foreach \r/\i in {0/2, 1.2/3, 2.4/4, 3.6/5, 4.8/6}
 {
 	\node  (P) at (\r, 0){};
 	\node[cell](a\i)[right=of P]{};
 	\path ([xshift=0cm,yshift=0.2cm]P.east) edge  ([xshift=0cm, yshift=0.2cm]a\i.west);
 	\path ([xshift=0cm,yshift=-0.2cm]a\i.west) edge  ([xshift=0cm, yshift=-0.2cm]P.east);	
 	 	
 }
 \draw[dashed, rounded corners](-0.7, -0.8)  rectangle (6.5, 0.8) ;
 \node[bar, minimum height = 0.2cm](p1)[above=of a1, yshift=-0.3cm]{};
 \node[bar, minimum height = 0.5cm](p2)[above=of a2, xshift=0.cm,yshift=-0.3cm]{};
 \node[bar, minimum height = 0.9cm](p3)[above=of a3, xshift=0.cm,yshift=-0.3cm]{};
 \node[bar, minimum height = 0.6cm](p4)[above=of a4, xshift=0.cm,yshift=-0.3cm]{};
 \node[bar, minimum height = 0.2cm](p5)[above=of a5, xshift=0.cm,yshift=-0.3cm]{};
 \node[bar, minimum height = 0.5cm](p6)[above=of a6, xshift=0.cm,yshift=-0.3cm]{};
 \node[vec, minimum height = 1.0cm](v1)[above=of p3, xshift=0.cm,yshift=0.0cm]{};
 \path[dashed,-] (p1.north) edge  (v1.south);
 \path[dashed,-] (p2.north) edge  (v1.south);
 \path[dashed,-] (p3.north) edge  (v1.south);
 \path[dashed,-] (p4.north) edge  (v1.south);
 \path[dashed,-] (p5.north) edge  (v1.south);
 \path[dashed,-] (p6.north) edge  (v1.south);
 \path[thick] (a1.north) edge  (p1.south);
 \path[thick] (a2.north) edge  (p2.south);
 \path[thick] (a3.north) edge  (p3.south);
 \path[thick] (a4.north) edge  (p4.south);
 \path[thick] (a5.north) edge  (p5.south);
 \path[thick] (a6.north) edge  (p6.south);
 \draw[ultra thick,-] (-0.7, 1.2) -- (6.5, 1.2);
 \node (t2) at (1.8, 3.6) {$\mathbf{h}_{s,1}^{u}$};
 
 \node[cell](b1) at (-0.1, 7.2) {};
 \draw (-2, 7.2) -- node[above]{A} (-0.7, 7.2);
 \foreach \r/\i in {0/2, 1.2/3, 2.4/4, 3.6/5, 4.8/6}
 {
    \node  (P) at (\r, 7.2){};
  	\node[cell](b\i)[right=of P]{};
  	\path ([xshift=0cm,yshift=0.2cm]P.east) edge  ([xshift=0cm, yshift=0.2cm]b\i.west);
  	\path ([xshift=0cm,yshift=-0.2cm]b\i.west) edge  ([xshift=0cm, yshift=-0.2cm]P.east);	
  	
 }
 \draw[dashed, rounded corners](-0.7, 6.4)  rectangle (6.5, 8) ;
 \node[bar, fill=red, minimum height = 0.5cm](h1)[below=of b1, yshift=0.3cm]{}; 
 \node[bar, fill=red, minimum height = 0.2cm](h2)[below=of b2, xshift=0cm, yshift=0.3cm]{}; 
 \node[bar, fill=red, minimum height = 0.4cm](h3)[below=of b3, xshift=0cm, yshift=0.3cm]{}; 
 \node[bar, fill=red, minimum height = 0.9cm](h4)[below=of b4, xshift=0cm, yshift=0.3cm]{};  
 \node[bar, fill=red, minimum height = 0.6cm](h5)[below=of b5, xshift=0cm, yshift=0.3cm]{}; 
 \node[bar, fill=red, minimum height = 0.3cm](h6)[below=of b6, xshift=0cm, yshift=0.3cm]{};  
 \node[vec, minimum height = 1.0cm](v2)[below=of h4, xshift=0.cm,yshift=0.05cm]{};
 \path[dashed,-] (h1.south) edge  (v2.north);  
 \path[dashed,-] (h2.south) edge  (v2.north); 
 \path[dashed,-] (h3.south) edge  (v2.north);  
 \path[dashed,-] (h4.south) edge  (v2.north); 
 \path[dashed,-] (h5.south) edge  (v2.north);  
 \path[dashed,-] (h6.south) edge  (v2.north); 
 \path[thick] (b1.south) edge  (h1.north); 
 \path[thick] (b2.south) edge  (h2.north);
 \path[thick] (b3.south) edge  (h3.north); 
 \path[thick] (b4.south) edge  (h4.north);
 \path[thick] (b5.south) edge  (h5.north); 
 \path[thick] (b6.south) edge  (h6.north);  
 \draw[ultra thick,-] (-0.7, 6) -- (6.5, 6);  
 \node (t4) at (3.0, 3.6) {$\mathbf{h}_{s,1}^{a}$};
 \draw[thick] (4, 3.6) -- (5, 3.6);
 \draw[dashed, rounded corners](5.2, 2.7)  rectangle (6.5, 4.5) ; 
 \draw[ultra thick,-] (5.3, 2.9) -- (5.3, 4.3);
 \node[vvec, minimum width =0.3cm]  (g) at (5.45, 4.0){};
 \node[vvec, minimum width =0.8cm]  (g) at (5.70, 3.6){};
 \node[vvec, minimum width =0.5cm]  (g) at (5.55, 3.2){};
 \node (t5) at (5.7, 2.4) {Slot Gate};
 
 \node[cell](d) at (10, 3.6) {};
 \draw[thick] (8.7, 3.6) -- node[above]{$\mathbf{x}_{s}^{0}$}  (9.7, 3.6);
 \node (t6) at (8, 4) {domain};
 \node (t7) at (8, 3.6) {$+$};
 \node (t8) at (8, 3.2) {slot};
 \draw[thick] (10, 2.6) -- node[right]{$\mathbf{c}_{s}$} (10, 3.1);
 \node (t8) at (10, 2.4) {slot-level context};
 \draw[thick] (10, 4.1) -- node[right]{$\mathbf{o}_{s}^{1}$} (10, 4.6);
 \node[cell, minimum width =1.3cm, minimum height = 0.5cm](pp1) at (12.6, 3.6) {\small$P_{s,1}^{v}$};
 \node (tt1) at (12.6, 4.2) {\small$\times\alpha_{s,1}$};
 \path[thick, ] (d.east) edge (pp1.west);
 \draw[thick] (11.5, 3.6) -- (11.5, 5.1) -- (6.4, 5.1) -- (6.4, 5.9);
 \draw[thick] (11.5, 3.6) -- (11.5, 2.1) -- (6.4, 2.1) -- (6.4, 1.3);
 \node[cell, minimum width =1.3cm, minimum height = 0.5cm](pp2) [above=of pp1, yshift=0.8cm] {\small$P_{s,1}^{a}$};
 \node (tt2) at (12.6, 6.6) {\small$\times(1-\alpha_{s,1})\beta_{s,1}$};
 \node[cell, minimum width =1.3cm, minimum height = 0.5cm](pp3) [below=of pp1, yshift=-0.8cm] {\small$P_{s,1}^{u}$};
 \node (tt3) at (12.6, 1.8) {\small$\times(1-\alpha_{s,1})(1-\beta_{s,1})$};
 \draw[thick] (6.4, 6) -- (11.95, 6);
 \draw[thick] (6.4, 1.2) -- (11.95, 1.2); 
 \draw[thick] (13.25, 6) -- (14.25, 6);
 \draw[thick] (13.25, 1.2) -- (14.25, 1.2);  
 \draw[thick] (13.25, 3.6) -- (14.25, 3.6); 
 \node[vvec, fill=green!20, minimum height = 0.3cm, minimum width =1cm]  (o1) at (14.78, 6.5){};
 \node[vvec, fill=green!20, minimum height = 0.3cm, minimum width =0.2cm]  (o2) [below=of o1, xshift=-0.375cm,yshift=0.8cm] {}; 
 \node[vvec, fill=green!20, minimum height = 0.3cm, minimum width =0.5cm]  (o3) [below=of o2, xshift=0.10cm,yshift=0.8cm] {}; 
 \node[vvec, fill=green!20, minimum height = 0.3cm, minimum width =0.8cm]  (o4) [below=of o3, xshift=0.15cm,yshift=0.8cm] {}; 
 \node[vvec, fill=green!20, minimum height = 0.3cm, minimum width =1.5cm]  (o5) [below=of o4, xshift=0.35cm,yshift=0.8cm] {};  
 \node[vvec, fill=green!20, minimum height = 0.3cm, minimum width =0.9cm]  (o6) [below=of o5, xshift=-0.3cm,yshift=0.8cm] {}; 
 \node[vvec, fill=green!20, minimum height = 0.3cm, minimum width =0.3cm]  (o7) [below=of o6, xshift=-0.3cm,yshift=0.8cm] {};
 \node[vvec, fill=green!20, minimum height = 0.3cm, minimum width =1.8cm]  (o8) [below=of o7, xshift=0.75cm,yshift=0.8cm] {};   
 \node[vvec, fill=green!20, minimum height = 0.3cm, minimum width =1.2cm]  (o9) [below=of o8, xshift=-0.3cm,yshift=0.8cm] {};  
 \node[vvec, fill=green!20, minimum height = 0.3cm, minimum width =0.2cm]  (o10) [below=of o9, xshift=-0.5cm,yshift=0.8cm] {}; 
 \node[vvec, fill=green!20, minimum height = 0.3cm, minimum width =0.7cm]  (o11) [below=of o10, xshift=0.25cm,yshift=0.8cm] {};  
 \node[vvec, fill=green!20, minimum height = 0.3cm, minimum width =1cm]  (o12) [below=of o11, xshift=0.15cm,yshift=0.8cm] {}; 
 \draw[ultra thick,-] (14.25, 0.2) -- (14.25, 7);
 \node (t7) at (17, 3.6) {$P_{s,1}$};
    
\end{tikzpicture}
			}
		\end{tabular}
	\end{center}
	\caption{The architecture of the Value Generator and the Slot Gate.}
	\label{fig:dec}
\end{figure*}
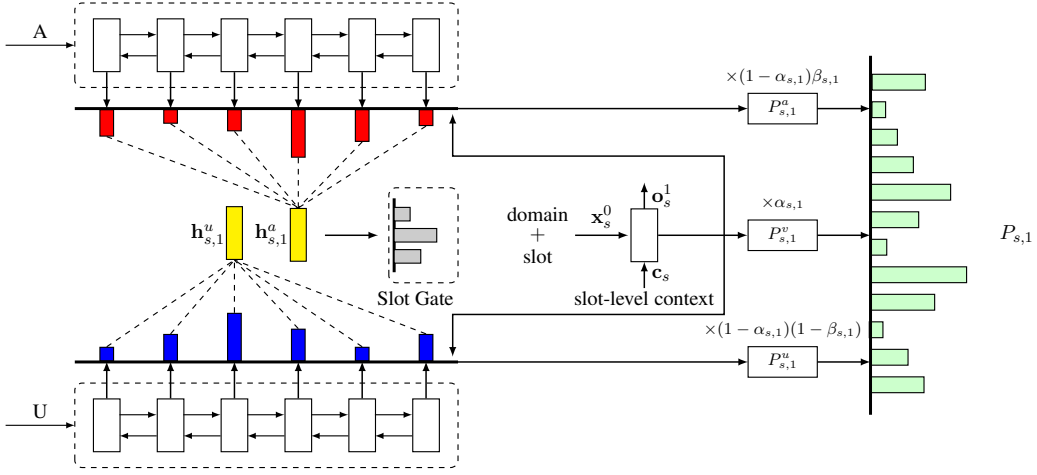
\subsection{Slot-level Context} \label{sec:localContext}
The purpose of applying the slot-level context here is to strengthen the context representation with slot specific features and deal with the slot overlapping problem. We employ the attention mechanism to construct the slot-level context. Specifically, for each $(domain, slot)$ pair, we introduce an embedding vector $\mathbf{v}_s$. The slot-level system context $\mathbf{c}_s^a$ and the slot-level user context $\mathbf{c}_s^u$ are computed by 
\begin{equation}
\begin{split}
\mathbf{c}_s^a \!= \!\sum_{i=1}^{M}\mu_{i}\mathbf{h}_{i}^{a}, \quad \!\!\mu_{i}\! =\! \frac{\exp{(\mathbf{v}_s^T}\mathbf{h}_{i}^a)}{\sum_{k=1}^{M}\exp{(\mathbf{v}_s^T}\mathbf{h}_{k}^a)}\\
\mathbf{c}_s^u\! =\! \sum_{j=1}^{N}\eta_{j}\mathbf{h}_{j}^{u}, \quad \!\!\eta_{j} \!=\! \frac{\exp{(\mathbf{v}_s^T}\mathbf{h}_{j}^u)}{\sum_{l=1}^{N}\exp{(\mathbf{v}_s^T}\mathbf{h}_{l}^u)}\\
\end{split}
\label{eq:context}
\end{equation}
The slot-level context of the entire dialogue history is then simply the summation of the slot-level system context and the slot-level user context
\begin{equation}
\begin{split}
\mathbf{c}_s = \mathbf{c}_s^a + \mathbf{c}_s^u
\end{split}
\label{eq:contextall}
\end{equation}
The slot-level context is then feed into the Value Generator as the initialized hidden state for the decoder GRU.

\subsection{Value Generator}
The Value Generator takes the slot-level context as input and uses a GRU decoder to generate the value sequence for each $(domain, slot)$ pair. Different from the copy mechanism applied in TRADE~\cite{Wu:19} that copying words from the entire dialogue history, in this paper, we propose a distributed copy mechanism that allows the state generator copying words from different sequences. The architecture of the Value Generator is shown in Figure~\ref*{fig:dec}. we now describe it in detail.


We use the abbreviation {\rm GRD} to denote the GRU decoder. At the $t^{th}$ decoding step, the hidden state of the GRU decoder for each $(domain, slot)$ pair $s$ is
\begin{equation}
\begin{split}
    \mathbf{o}_s^t = {\rm GRD}(\mathbf{x}_s^{t}, \mathbf{o}_s^{t-1}, \mathbf{W}_{d})
\end{split}
\label{eq:dec}
\end{equation}
where $\mathbf{x}_{s}^{t}$ is the input at the $t^{th}$ step, $\mathbf{o}^{t}_{s}$ is the hidden state at the $t^{th}$ step and $\mathbf{W}_{d}$ is the parameters of the GRU decoder. The hidden state of GRD for each slot is initialized with corresponding slot-level context $\mathbf{c}_{s}$. The first input $\mathbf{x}_{s}^{0}$ is set as the summation of corresponding domain embedding and slot embedding.

We then introduce three distributions on the vocabulary: $P_{s,t}^{v}$, $P_{s,t}^{a}$ and $P_{s,t}^{u}$, for applying distributed copy mechanism. The three distributions represent the 
probabilities of generating a word from the vocabulary, copying a word from the historical system utterances and copying a word from the historical user utterances, respectively.
Let $\mathbf{e}_{i}$ be the embedding of the $i^{th}$ word in the vocabulary and $|V|$ be the vocabulary size. We use $P_{s,t}[i]$ to denote the $i^{th}$ element in $P_{s,t}$. Then the three distributions are computed by
\begin{equation}
\begin{split}
&P_{s,t}^{v}[i] = \frac{\exp{(\mathbf{e}_{i}^{T}\mathbf{o}_s^t)}}{\sum_{j=1}^{|V|}\exp{(\mathbf{e}_{j}^{T}\mathbf{o}_s^t)}}\\
&P_{s,t}^{a}[i] = \sum_{f(k)=i} \frac{\exp{((\mathbf{h}_{k}^{a})^{T}\mathbf{o}_s^t)}}{\sum_{j=1}^{M}\exp{((\mathbf{h}_{j}^{a})^{T}\mathbf{o}_s^t)}}\\ 
&P_{s,t}^{u}[i] = \sum_{f(k)=i} \frac{\exp{((\mathbf{h}_{k}^{u})^{T}\mathbf{o}_s^t)}}{\sum_{j=1}^{N}\exp{((\mathbf{h}_{j}^{u})^{T}\mathbf{o}_s^t)}}  
\end{split}
\label{eq:gen}
\end{equation}
where the function $f$ is used for mapping a distribution on the dialogue-history to corresponding distribution on the vocabulary. 


The three distributions, $P_{s,t}^{v}$, $P_{s,t}^{a}$ and $P_{s,t}^{u}$ are then combined by learnable weights. We define $\alpha_{s,t}$ as the weight of generating from the vocabulary and $\beta_{s,t
}$ as the weight of choosing to copy a word from the system utterances. For calculating the weights $\alpha_{s,t}$ and $\beta_{s,t}$, we first generate new feature vectors 
\begin{equation}
\begin{split}
\mathbf{h}_{s,t}^a \!= \!\sum_{i=1}^{M}q_{s,t}^{a}[i]\mathbf{h}_{i}^{a}, \quad \!\!q_{s,t}^{a}[i]\! =\! \frac{\exp{(\mathbf{o}_s^t}\cdot\mathbf{h}_{i}^a)}{\sum_{k=1}^{M}\exp{(\mathbf{o}_s^t}\cdot\mathbf{h}_{k}^a)}\\
\mathbf{h}_{s,t}^u\! =\! \sum_{j=1}^{N}q_{s,t}^{a}[j]\mathbf{h}_{j}^{u}, \quad \!\!q_{s,t}^{a}[j] \!=\! \frac{\exp{(\mathbf{o}_s^t}\cdot\mathbf{h}_{j}^u)}{\sum_{l=1}^{N}\exp{(\mathbf{o}_s^t}\cdot\mathbf{h}_{l}^u)}\\
\end{split}
\label{eq:h_feature}
\end{equation}
The weight $\alpha_{s,t}$ and $\beta_{s,t}$ are then computed by
\begin{equation}
\begin{split}
    &\alpha_{s,t} = \sigma(\mathbf{W}_{v}^{T}\cdot [\mathbf{x}_s^t, \mathbf{o}_s^t, \mathbf{h}_{s,t}^{a}, \mathbf{h}_{s,t}^{u}])\\
	&\rho_{s,t}^{a} = \mathbf{W}_{c}^{T}\cdot [\mathbf{x}_s^t, \mathbf{o}_s^t, \mathbf{h}_{s,t}^{a}]\\
	&\rho_{s,t}^{u} = \mathbf{W}_{c}^{T}\cdot [\mathbf{x}_s^t, \mathbf{o}_s^t, \mathbf{h}_{s,t}^{u}]\\
	&\beta_{s,t} = \frac{\exp{(\rho_{s,t}^{a})}}{\exp{(\rho_{s,t}^{a})}+\exp{(\rho_{s,t}^{u})}}        
\end{split}
\label{eq:vocweight}
\end{equation}
where $\mathbf{W}_{v}$ and $\mathbf{W}_{c}$ are the parameters of the linear functions, and $\sigma$ denotes the logistic function.

%

The final distribution $P_{s,t}$ is then calculated as the weighted sum of distributions $P_{s,t}^{v}$, $P_{s,t}^{a}$ and $P_{s,t}^{u}$ as follows
\begin{equation}
\begin{split}
    P_{\!\!s,t} \!=\! \alpha_{s,t}P_{\!\!s,t}^{v} \!+\! (1\!-\!\alpha_{s,t})(\beta_{s,t}P_{\!\!s,t}^{a} \!+\! (1\!-\!\beta_{s,t})P_{\!\!s,t}^{u})
\end{split}
\label{eq:final}
\end{equation}

The $t^{th}$ word of the value for $(domain,slot)$ pair $s$ is then generated from distribution $P_{s,t}$. The embedding of the generated word is then used as the next input of the GRU decoder. This generation procedure allows the state generator to generate words from the vocabulary or copy words from either the historical system utterances or the historical user utterances. 

\subsection{Slot Gate}
Following TRADE~\cite{Wu:19}, we introduce the slot gate to predict the special values {\em none} (the value of the slot is not expressed yet) and {\em dontcare} (the user does not care about the slot) for each $(domain, slot)$ pair. Specifically, the slot gate is a three-class classifier, which aims to identify whether the value {\em none}, {\em dontcare} or other value is expressed from the context through a softmax classifier
\begin{equation}
\begin{split}
P_s^c = \softmax(\mathbf{W}_s^T \cdot [\mathbf{h}_{s,1}^{a}, \mathbf{h}_{s,1}^{u}])
\end{split}
\label{eq:psc}
\end{equation}
where $\mathbf{W}_s$ is the parameter of the softmax classifier. For a $(domain, slot)$ pair, if the output of the slot gate is {\em none} or {\em dontcare}, the generated word sequence from the state generator will be ignored and the corresponding predicted result of the slot gate will be chosen as the value. Otherwise, the generated word sequence from the state generator will be the predicted value for the $(domain, slot)$ pair.

\subsection{Loss Function and Optimization}
The cross-entropy loss is built for optimizing both the Value Generator and the Slot Gate, simultaneously. Let $\mathcal{S}$ be the total set of $(doamin, slot)$ pairs, and $T_{s}$ be the number of words in the value for slot $s\in \mathcal{S}$. We define $\mathbf{y}_{s}^{c}$ as the ground-truth one-hot label vector of the slot gate and $\mathbf{y}_{s,t}^{v}$ as the one-hot representation of the $t^{th}$ word in the value of $s$. The loss function is then defined as 
\begin{equation}
\begin{split}
    \mathcal{L} =& \sum_{s \in \mathcal{S}}\sum_{i=1}^{3}-\mathbf{y}_{s}^{c}[i]\cdot \log P_{s}^{c}[i] \\
    +& \sum_{s \in \mathcal{S}} \sum_{t=1}^{T_s} \sum_{j=1}^{|V|} -\mathbf{y}_{s,t}^{v}[j]\cdot \log P_{s,t}[j]
\end{split}
\label{eq:loss}
\end{equation}
The loss function can be optimized by stochastic gradient descent(SGD) method.
\section{Experiment}
\subsection{Datasets}
\noindent {\bf MultiWOZ 2.0.} The Multi-Domain Wizard-of-Oz (MultiWOZ 2.0) dataset, collected by ~\cite{Budzianowski:18}, with conversations spanning over multiple domains and topics, is used to train and evaluate the models.
There are total 7 domains with 30 $(domain, slot)$ pairs in the ontology; these $(domain, slot)$ pairs involve $4,510$ values. The dataset contains $10,419$ dialogues with a a total $115,434$ turns; the average turns of dialogue is $13.46$.
The training, validation and test set contain $8,420$, $1,000$ and $1,000$ dialogues respectively.
As is mentioned in ~\cite{Wu:19} that {\em hospital} and {\em police} domain has very few dialogues and only appear in the training set. We thus follow the dataset setting in ~\cite{Wu:19} that only keep five domains ({\em restaurant}, {\em hotel}, {\em attraction}, {\em taxi}, {\em train}) in the experiment.

\noindent {\bf MultiWOZ 2.1.} As is pointed out in~\cite{Eric:19}, the MultiWOZ 2.0 dataset is faulty in substantial errors in the state annotations and dialogue utterances. In order to clean the dataset, the authors of ~\cite{Eric:19} ask crowd-source workers to fix the state annotations and utterances in the original data. As a result, over 32\% of state annotations in 40\% of the dialogue turns are changed and 146 utterances are fixed. The cleaned dataset is released as the MultiWOZ 2.1 dataset. We also evaluate our models on the MultiWOZ 2.1 dataset.

\subsection{Implementation Details}
The models are implemented using the Pytorch framework. The code and data are released on the Github page\footnote{https://github.com/BDBC-KG-NLP/PIN\_EMNLP2020}. The word embeddings are initialized by the concatenation of the pre-trained GloVe embeddings~\cite{Pennington:14} and character n-gram embeddings~\cite{Hashimoto:17}. 
The batch size is set as 32. The dimensions of hidden states in GRUs are set as $400$. The embedding dropout is used in the Interactive Encoder with a dropout rate $0.3$.Following~\cite{Bowman:16, Wu:19}, we also adopt the word dropout in the Interactive Encoder to improve the model generalization; and the dropout rate is set as $0.3$. At training time, the Value Generator uses Teacher-forcing~\cite{Williams:89} with a probability $0.5$. The greedy search~\cite{Vinyals:15} is used in the decoding process. The Adam optimizer~\cite{Kingma:15} with an initialized learning rate $0.001$ to optimize the model.

\subsection{Evaluation Metrics} 
The standard metrics joint goal accuracy and goal accuracy are used to evaluate the multi-domain dialogue state tracking performance. The joint goal accuracy denotes the proportion of dialogue turns where the values of all the $(domain, slot)$ pairs are correctly predicted. While goal accuracy is the proportion of slots whose values are correctly predicted. 

\subsection{Baseline Models}
The recently proposed dialogue state tracking models are used for comparison. 
The models dealing with dialogue state tracking through building classifiers on predefined ontology include the MDBT~\cite{Ramadan:18}, GLAD~\cite{Zhong:18}, GCE~\cite{Nouri:18}, SUMBT~\cite{Lee:19}, FJST~\cite{Eric:19}, HJST~\cite{Eric:19} and SST~\cite{Chen:20}. The models utilizing the copy system include PtrNet~\cite{Xu:18}. The models incorporating both classifiers and copy system include HyST~\cite{Goel:19}, DSTreader~\cite{Gao:19}, TRADE~\cite{Wu:19}, DST-Picklist~\cite{Zhang:19} and MERET~\cite{Huang:20}.

To investigate how much the proposed interactive encoder and distributed copy mechanism contributes to the PIN model, we also report the results of two ablated version of the PIN model: PIN--inter and PIN--dcopy. The PIN--inter model removes the interaction between the two parallel encoders in PIN and allows them to be independent. And the PIN--dcopy model copies words from the entire dialogue history instead of applying the distributed copy.  

\begin{table}[ht]
	\centering
	\caption{Evaluation on the MultiWOZ 2.0 dataset.}
	\label{tab:multiWOZ2.0}
	\begin{tabular}{ccc}
		\toprule
		{\bf Model}&{\bf Joint Goal} (\%) & {\bf Goal} (\%) \\
		\midrule
		MDBT& 15.57 & 89.53\\
		PtrNet & 30.28 & 93.85\\
		GLAD & 35.57 & 95.44\\
		GCE & 36.27 & {\bf 98.42}\\
		HJST & 38.40 & -\\
		DSTreader & 39.41 & -\\
		FJST & 40.20 & -\\
		HyST & 42.33 & -\\
		HyST(ensemble) & 44.22 & -\\		
		SUMBT & 42.40 & -\\
		DSTreader+JST & 47.33 & -\\		
		TRADE & 48.62 & 96.92\\
		MERET & 50.91 & 97.07 \\
		SST & 51.17 & - \\
		{\bf PIN--inter} & 51.95 & 97.24\\
		{\bf PIN--dcopy} & 50.57 & 97.06\\		
		{\bf PIN} & {\bf 52.44} & 97.28\\
		\bottomrule
	\end{tabular}
\end{table}

\subsection{Experimental Results}
\noindent {\bf Evaluation on the MultiWOZ 2.0 dataset.} The evaluation results on the MultiWOZ 2.0 dataset are shown in Table ~\ref{tab:multiWOZ2.0}. We observe that most of the models building classifiers and the models using the copy system to generate the states are inferior to the models utilizing both the classifiers and the copy system. As mentioned in ~\cite{Eric:19}, the models building upon a copy system have an advantage in extracting values from the dialogue history but struggle to predict values that do not exist in the dialogue history. Thus it is reasonable that models combining copy systems with state classifiers achieve better performance. Compared with the baseline model TRADE and the previous state-of-the-art model SST, PIN achieves significant 3.82\% and 1.27\% performance gain. This fact demonstrates that the modeling of the interaction dependencies, the slot-level context and the distributed copy mechanism help improve state generation.
\begin{table}[ht]
  \centering
  \caption{Evaluation on the MultiWOZ 2.1 dataset.}
  \label{tab:multiWOZ2.1}
  \begin{tabular}{ccc}
    \toprule
    {\bf Model}&{\bf Joint Goal} (\%) & {\bf Goal} (\%) \\
    \midrule
    HJST & 35.55 & -\\
    DST Reader & 36.40 & -\\
    FJST & 38.00 & -\\
    HyST & 38.10 & -\\
    TRADE & 45.60 & 96.55\\
    DST-Picklist & 53.30 & -\\
    SST & {\bf 55.23} & - \\
    {\bf PIN--inter} & 47.36 & 96.90\\
    {\bf PIN--dcopy} & 47.29 & 96.91\\
    {\bf PIN} & 48.40 & {\bf 97.02}\\
  \bottomrule
\end{tabular}
\end{table}

\noindent {\bf Evaluation on the MultiWOZ 2.1 dataset.} The evaluation results on the MultiWOZ 2.1 dataset are shown in Table ~\ref{tab:multiWOZ2.1}. The consistent performance drop is caused by changing a value to a {\em dontcare} or {\em none} label as explained in ~\cite{Eric:19}. The PIN model outperforms the previous models except for the DST-Picklist and SST model, which indicates the effectiveness of the model design. Although DST-Picklist and SST achieve better performance than PIN, DST-Picklist takes a lot of human efforts in dividing the slots into span-based or picklist-based slots and SST requires extra relation information among the slots. PIN's performance drop in the ablated version (PIN-inter and PIN-dcopy)
on both datasets demonstrates the necessity of encoder-interaction and distributed copy. 
\begin{table}[ht]
	\centering
	\caption{The evaluation results of overlapping slots and non-overlapping slots on the MultiWOZ 2.1 dataset. 1:{\em Restaurant}, 2:{\em Hotel}, 3:{\em Attraction}, 4:{\em Train}, 5:{\em Taxi}.}
	\label{tab:overlap}
	\begin{tabular}{cccc}
		\toprule
		{\bf Slot} & {\bf Domains} & {\bf TRADE} & {\bf PIN}\\
		\midrule
		area& 1,2,3 & 86.2 & {\bf 86.4} \\
		book people& 1,2,3 & 92.0 & {\bf 95.1}\\	
		price range& 2,3 & 84.2 & {\bf 89.7}\\
		book day& 2,3 & 96.4 & {\bf 96.8}\\
		departure& 4,5 & 89.0 & {\bf 90.9}\\    
		destination& 4,5 & 91.6 & {\bf 92.4}\\ 
		leave at& 4,5 & 65.1 & {\bf 66.7}\\    
		arrive by& 4,5 & 82.4 & {\bf 84.7}\\		 book time& 1 & {\bf 92.7} & 91.8 \\
		food & 1 & {\bf 92.8} & 92.6 \\		
		parking & 2 & 80.1 & {\bf 81.2} \\
		book stay & 2 & {\bf 96.4} & 96.2 \\
		internet & 2 & 78.8 & {\bf 81.2} \\
		\bottomrule
	\end{tabular}
\end{table}
\subsection{Evaluation on the Overlapping Slots} 
In multi-domain dialogue state tracking, domains may have overlapping slots. One of the motivations for building the PIN model is to handle the slot overlapping problem with a slot-level context. Thus we report the goal accuracy on overlapping slots and non-overlapping slots in Table~\ref{tab:overlap} for further analysis on PIN. Table~\ref{tab:overlap} shows that slot overlapping (involve at least two domains) usually appears among similar domains, such as ({\em Restaurant}, {\em Hotel}, {\em Attraction}) and ({\em Train}, {\em Taxi}). The PIN model achieves much higher goal accuracy than TRADE on all overlapping slots, compared with non-overlapping slots. This result demonstrates the effectiveness of the slot-level context on extracting distinctive features for each slot so that the values for overlapping slots are correctly predicted. 
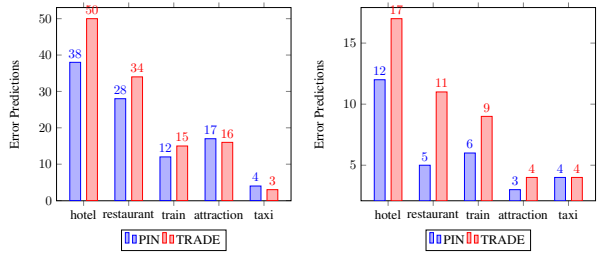
\begin{figure}[ht]
	\begin{center}
		\begin{tabular}{lr}
		\hspace{-.4cm}
			\scalebox{0.45}{
				\begin{tikzpicture}
\begin{axis}[
	x tick label style={
		/pgf/number format/1000 sep=},
	ylabel=Error Predictions,
	enlarge x limits=0.15,
	enlarge y limits=0.065,
	legend style={at={(0.5,-0.15)},
		anchor=north,legend columns=-1},
	ybar=5pt,
	bar width=9pt,
	symbolic x coords={hotel,restaurant,train,attraction,taxi},
	nodes near coords,
]
\addplot 
	coordinates {(hotel,38) (restaurant,28)
		  (train,12) (attraction,17) (taxi,4)};

\addplot 
	coordinates {(hotel,50) (restaurant,34)
		(train,15) (attraction,16) (taxi,3)};

\legend{PIN,TRADE}
\end{axis}
\end{tikzpicture}
			}&
		\hspace{-.4cm}
			\scalebox{0.45}{
				\begin{tikzpicture}
\begin{axis}[
	x tick label style={
		/pgf/number format/1000 sep=},
	ylabel=Error Predictions,
	enlarge x limits=0.15,
	enlarge y limits=0.065,
	legend style={at={(0.5,-0.15)},
		anchor=north,legend columns=-1},
	ybar=5pt,
	bar width=9pt,
	symbolic x coords={hotel,restaurant,train,attraction,taxi},
	nodes near coords,
]
\addplot 
	coordinates {(hotel,12) (restaurant,5)
		  (train,6) (attraction,3) (taxi,4)};

\addplot 
	coordinates {(hotel,17) (restaurant,11)
		(train,9) (attraction,4) (taxi,4)};

\legend{PIN,TRADE}
\end{axis}
\end{tikzpicture}
			}			
		\end{tabular}
	\end{center}
	\caption{Error analysis on the dialogue turns involving in-turn dependencies (left) and cross-turn dependencies (right). We report the number of error predictions for TRADE and PIN on each domain.}
	\label{fig:bar}
\end{figure}
\subsection{The Effectiveness of the Interactive Encoder}\label{sec:sigleDomain}
To study the Interactive Encoder in handling in-turn and cross-turn dependencies in MDST, we make a error analysis on a subset of the test data. We first sample 100 dialogue turns from the MultiWOZ 2.1 test set. Then the wrongly predicted $(domain, slot, value)$ triplets for TRADE and PIN are selected and each of the triplets is marked according to the dependencies (in-turn or cross-turn) involved. The statistics of these error predictions are shown in Figure ~\ref{fig:bar}. We observe that whether in the dialogue turns involving in-turn dependencies or cross-turn dependencies, the PIN model creates much fewer prediction errors than the TRADE model, especially on $hotel$ domain and $restaurant$ domain.
These results demonstrate the effectiveness of the Interactive Encoder in capturing the in-turn and cross-turn dependencies.
\begin{figure*}[ht]
	\begin{center}
		\begin{tabular}{c}
			\begin{tikzpicture}[-latex ,auto ,node distance =0.0 cm and 0.0 cm, on grid ,
    	semithick ,				
    	box/.style ={rectangle, draw=white, fill=red, text=black, minimum width =0.1cm, minimum height = 0.1cm}]
				
		\node[rectangle, draw=black, minimum width =14cm, minimum height=5cm](bigBox){};
		\node[rectangle, draw=black, minimum width =14cm, minimum height=1cm, above= of bigBox, yshift=2.0cm](title){{\bf Domain: Restaurant} \hspace{0.5cm} {\bf Slot: food} \hspace{0.5cm} {\bf Value: european} \hspace{0.5cm} $\mathbf{\alpha=0.37}$};
        \node[rectangle, align=center, draw=black, minimum width =4cm, minimum height=2cm, above= of bigBox, xshift=-5cm, yshift=0.5cm](sysTitle){\parbox{3cm}{\centering System Utterances \\ \vspace{0.2cm} $\beta=0.033$}};	
        \node[rectangle, draw=black, minimum width =4cm, minimum height=2cm, above= of bigBox, xshift=-5cm, yshift=-1.5cm](usrTitle){\parbox{3cm}{\centering User Utterances \\ \vspace{0.2cm} $1-\beta=0.967$}};  
        \node[rectangle, align=center, draw=black, minimum width =10cm, minimum height=2cm, above= of bigBox, xshift=2cm, yshift=0.5cm](sysText){};  
        \node[rectangle, align=center, draw=black, minimum width =10cm, minimum height=2cm, above= of bigBox, xshift=2cm, yshift=-1.5cm](usrText){};

        \node[box, fill opacity=0.0, text opacity=1.0, xshift=-2.2cm, yshift=1.0cm ](s0){PAD ;}; 
        \node[box, fill opacity=0.00015, text opacity=1.0, right= of s0, xshift=0.8cm, yshift=0.05cm](s1){ok};   
        \node[box, fill opacity=0.00, text opacity=1.0, right= of s1, xshift=0.4cm](s2){,};   
        \node[box, fill opacity=0.026, text opacity=1.0, right= of s2, xshift=0.4cm](s3){i};                                  
        \node[box, fill opacity=0.14228, text opacity=1.0, right= of s3, xshift=0.7cm](s4){found};  
        \node[box, fill opacity=0.0001, text opacity=1.0, right= of s4, xshift=0.9cm](s5){the};  
        \node[box, fill opacity=0.000, text opacity=1.0, right= of s5, xshift=3.1cm, yshift=-0.03cm](s6){cambridge lodge restaurant . would};  
        \node[box, fill opacity=0.0001, text opacity=1.0, below= of s0, xshift=-0.12cm, yshift=-0.6cm](s6){you};  
        \node[box, fill opacity=0.00, text opacity=1.0, right= of s6, xshift=0.7cm, yshift=0.02cm](s7){like};   
        \node[box, fill opacity=0.00, text opacity=1.0, right= of s7, xshift=1.7cm](s8){$\cdots \cdots$\hspace{0.1cm} i \hspace{-0.1cm} would};   
        \node[box, fill opacity=0.0023, text opacity=1.0, right= of s8, xshift=3.8cm](s9){suggest \hspace{-0.1cm} $\cdots \cdots$ . \hspace{-0.1cm} would you like};   
        \node[box, fill opacity=0.0164, text opacity=1.0, right= of s9, xshift=2.8cm](s10){me};    
        \node[box, fill opacity=0.0001, text opacity=1.0, below= of s6, xshift=0.56cm, yshift=-0.5cm](s11){to make a};    
        \node[box, fill opacity=0.668, text opacity=1.0, right= of s11, xshift=1.8cm](s12){reservation};  
        \node[box, fill opacity=0.038, text opacity=1.0, right= of s12, xshift=1.2cm](s13){?};   
        \node[box, fill opacity=0.00, text opacity=1.0, right= of s12, xshift=2.0cm](s14){$\cdots \cdots$}; 
        
        \node[box, fill opacity=0.0, text opacity=1.0, xshift=-1.15cm, yshift=-1.0cm ](s14){i am looking for a}; 
        \node[box, fill opacity=0.5069, text opacity=1.0, right= of s14, xshift=2.3cm, yshift=-0.06cm](s15){european}; 
        \node[box, fill opacity=0.08425, text opacity=1.0, right= of s15, xshift=1.7cm, yshift=0.07cm](s16){restaurant};         
        \node[box, fill opacity=0.00, text opacity=1.0, right= of s16, xshift=2.2cm](s17){in \hspace{0.05cm} the \hspace{0.05cm} west \hspace{0.05cm} of};                                                                                                               			
        \node[box, fill opacity=0.000, text opacity=1.0, below= of s14, xshift=0.0cm, yshift=-0.6cm](s18){cambridge; $\cdots \cdots$};
        \node[box, fill opacity=0.000, text opacity=1.0, right= of s18, xshift=4.2cm](s19){i really need someplace expensive ,}; 
        \node[box, fill opacity=0.000, text opacity=1.0, right= of s19, xshift=3.0cm](s20){it}; 
        \node[box, fill opacity=0.0001, text opacity=1.0, below= of s18, xshift=-1.19cm, yshift=-0.5cm](s21){is a};  
        \node[box, fill opacity=0.1051, text opacity=1.0, right= of s21, xshift=1.0cm](s22){special};  
        \node[box, fill opacity=0.2777, text opacity=1.0, right= of s22, xshift=1.4cm](s23){occasion};  
        \node[box, fill opacity=0.00, text opacity=1.0, right= of s23, xshift=1.9cm](s24){for me $\cdots \cdots$};                                          				
\end{tikzpicture}
		\end{tabular}
	\end{center}
	\caption{An example of dialogues and prediction of PIN. The red color represent the copy probability of the word. And the copy probability of the word {\em reservation} in system utterances is $0.668$, the copy probability of the word {\em european} in the user utterances is $0.507$.}
	\label{fig:case}
\end{figure*}
\subsection{The Function of the Distributed Copy Mechanism}\label{sec:case}
Unlike the traditional copy mechanism that only copies words from one sequence, the distributed copy mechanism in the PIN model can copy words from two separate sequences considering the interactive nature of dialogues. 
The example in Figure~\ref{fig:case} shows a case that the traditional copy mechanism will make a wrong prediction, but the distributed copy mechanism will correctly predict. The dialogue in Figure~\ref{fig:case} is a sample from the Restaurant domain in the test set. In this example, we want to predict the value of the {\em food} slot. As the wight $\alpha=0.37$, the generator has a higher probability of copying a word from the dialogue history. In the total dialogue history, if we ignore the wight $\beta$, which determines whether to copy from the historical system utterance or the historical user utterance, the generator will copy the wrong word {\em reservation} from the entire dialogue history because the word {\em reservation} has higher copy probability $0.668$ than $0.507$ of the word {\em european}. This wrong prediction will happen in the traditional copy-based model. But in PIN, the word to be copied also depends on the sequence-selection weight $\beta$. With a probability $0.967$ to copy the word from the historical user utterance, the correct value {\em european} will be copied according to Equation~\ref{eq:final}. This case demonstrates the effectiveness of the distributed copy mechanism.

\section{Related Works}
The dialogue state tracking (DST) problem has attracted the research community for years. The traditional DST models focus on single domain dialogue state tracking
~\cite{Thomson:10,Wang:13,Lee:16,Liu:17a,Jang:16,Shi:16,Vodol:17,Yu:15,Henderson:14,Zilka:15,Mrksic:17,Xu:18,Zhong:18,Ren:18}. Some of these models solve DST problem by incorporating a natural language understanding (NLU) module~\cite{Thomson:10,Wang:13} or jointly modeling NLU and DST~\cite{Henderson:14,Zilka:15}, which rely on hand-crafted features or delexicalisation features.
Other models adopt the representation learning approach and incorporate neural networks to extract features and track the dialogue states (NBT~\cite{Mrksic:17}, GLAD~\cite{Zhong:18}, StateNet~\cite{Ren:18}, PtrNet~\cite{Xu:18} and SUMBT~\cite{Lee:19}). Although these models have achieved remarkable success in single-domain DST , they can not be capable enough in multi-domain DST.

Recently, the multi-domain DST attracts more attention than the single-domain DST in the research community. The first work involving state tracking in multiple domains is ~\cite{Mrksic:15}. This work proposes a pre-training procedure to improve the performance on a new domain. The work of ~\cite{Rastogi:17} uses bi-directional GRU to extract features and predict the value by a candidate scoring model. The MDBT~\cite{Ramadan:18} model applies multiple bi-directional-LSTM to jointly track the domain and states. It adopts semantic similarity between the ontology and utterances and allows parameter sharing across domains. The HyST~\cite{Goel:19} model combines a classification-based system and an n-gram copy-based system to deal with multi-domain dialogue state tracking problem. The FJST and HJST model presented in~\cite{Eric:19} employ flatten structured LSTM and hierarchical structured LSTM to encode the dialogue history respectively. The TRADE model~\cite{Wu:19} combines the soft-copy mechanism to generate states and a slot gate to classify special values for each slot. These models motivate our design of the PIN model.

Another idea related to our design of PIN is hierarchical recurrent networks. The hierarchical recurrent networks have been used for dialogue representation in HRED~\cite{Serban:16} and VHRED~\cite{Serban:17}. Although our model has a slight flavor of a hierarchical structure (since a sentence-level encoding is sent to another GRU as its initial state), our model is very different from the hierarchical recurrent networks. Specifically, in PIN, the inputs and outputs for each GRU layer are both at the word level; and the GRU layers are parallel, albeit interacting. This is distinct from the hierarchical recurrent networks, where a GRU layer takes word-level inputs, and outputs at the sentence level; then the sentence-level representations are used as the inputs to the next GRU layer.
\section{Conclusion}
This paper studies the problem of state generation for multi-domain dialogues. Existing generation-based models fail to model the dialogue dependencies and ignore the slot-overlapping problem in MDST. To overcome the limitation of existing models, we present novel Parallel Interactive Networks (PIN) for more accurate and robust dialogue state generation. The design of the PIN model is inspired by the interactive nature of the dialogues and the overlapping slots in the ontology. The Interactive Encoder characterizes the cross-turn dependencies and the in-turn dependencies. The slot-overlapping problem is solved by introducing the slot-level context. Furthermore, a distributed copy mechanism is introduced to perform a selective copy from either the historical system utterances or the historical user utterances. Empirical studies on two benchmark datasets demonstrate the effectiveness of the PIN model.

\section*{Acknowledgment}
This work is supported partly by the National Natural Science Foundation of China (No. 61772059, 61421003), by the Beijing Advanced Innovation Center for Big Data and Brain Computing (BDBC), by the Fundamental Research Funds for the Central Universities, by the Beijing S\&T Committee (No. Z191100008619007) and by the State Key Laboratory of Software Development Environment (No. SKLSDE-2020ZX-14).
\bibliographystyle{acl_natbib}
\bibliography{emnlp2020}

\begin{thebibliography}{37}
\expandafter\ifx\csname natexlab\endcsname\relax\def\natexlab#1{#1}\fi

\bibitem[{Bowman et~al.(2016)Bowman, Vilnis, Vinyals, Dai, J{\'{o}}zefowicz,
  and Bengio}]{Bowman:16}
Samuel~R. Bowman, Luke Vilnis, Oriol Vinyals, Andrew~M. Dai, Rafal
  J{\'{o}}zefowicz, and Samy Bengio. 2016.
\newblock Generating sentences from a continuous space.
\newblock In \emph{Proceedings of the 20th {SIGNLL} Conference on Computational
  Natural Language Learning, CoNLL 2016, Berlin, Germany, August 11-12, 2016},
  pages 10--21.

\bibitem[{Budzianowski et~al.(2018)Budzianowski, Wen, Tseng, Casanueva, Ultes,
  Ramadan, and Gasic}]{Budzianowski:18}
Pawel Budzianowski, Tsung{-}Hsien Wen, Bo{-}Hsiang Tseng, I{\~{n}}igo
  Casanueva, Stefan Ultes, Osman Ramadan, and Milica Gasic. 2018.
\newblock Multiwoz - {A} large-scale multi-domain wizard-of-oz dataset for
  task-oriented dialogue modelling.
\newblock In \emph{Proceedings of the 2018 Conference on Empirical Methods in
  Natural Language Processing, Brussels, Belgium, October 31 - November 4,
  2018}, pages 5016--5026.

\bibitem[{Chen et~al.(2020)Chen, Lv, Wang, Zhu, Tan, and Yu}]{Chen:20}
Lu~Chen, Boer Lv, Chi Wang, Su~Zhu, Bowen Tan, and Kai Yu. 2020.
\newblock Schema-guided multi-domain dialogue state tracking with graph
  attention neural networks.
\newblock In \emph{The Thirty-Fourth {AAAI} Conference on Artificial
  Intelligence, {AAAI} 2020, The Thirty-Second Innovative Applications of
  Artificial Intelligence Conference, {IAAI} 2020, The Tenth {AAAI} Symposium
  on Educational Advances in Artificial Intelligence, {EAAI} 2020, New York,
  NY, USA, February 7-12, 2020}, pages 7521--7528. {AAAI} Press.

\bibitem[{Chung et~al.(2014)Chung, G{\"{u}}l{\c{c}}ehre, Cho, and
  Bengio}]{Chung:14}
Junyoung Chung, {\c{C}}aglar G{\"{u}}l{\c{c}}ehre, KyungHyun Cho, and Yoshua
  Bengio. 2014.
\newblock Empirical evaluation of gated recurrent neural networks on sequence
  modeling.
\newblock \emph{CoRR}, abs/1412.3555.

\bibitem[{Eric et~al.(2020)Eric, Goel, Paul, Sethi, Agarwal, Gao, Kumar, Goyal,
  Ku, and Hakkani{-}T{\"{u}}r}]{Eric:19}
Mihail Eric, Rahul Goel, Shachi Paul, Abhishek Sethi, Sanchit Agarwal, Shuyang
  Gao, Adarsh Kumar, Anuj~Kumar Goyal, Peter Ku, and Dilek Hakkani{-}T{\"{u}}r.
  2020.
\newblock Multiwoz 2.1: {A} consolidated multi-domain dialogue dataset with
  state corrections and state tracking baselines.
\newblock In \emph{Proceedings of The 12th Language Resources and Evaluation
  Conference, {LREC} 2020, Marseille, France, May 11-16, 2020}, pages 422--428.
  European Language Resources Association.

\bibitem[{Gao et~al.(2019)Gao, Sethi, Agarwal, Chung, and
  Hakkani{-}T{\"{u}}r}]{Gao:19}
Shuyang Gao, Abhishek Sethi, Sanchit Agarwal, Tagyoung Chung, and Dilek
  Hakkani{-}T{\"{u}}r. 2019.
\newblock Dialog state tracking: {A} neural reading comprehension approach.
\newblock In \emph{Proceedings of the 20th Annual SIGdial Meeting on Discourse
  and Dialogue, SIGdial 2019, Stockholm, Sweden, September 11-13, 2019}, pages
  264--273. Association for Computational Linguistics.

\bibitem[{Goel et~al.(2019)Goel, Paul, and Hakkani{-}T{\"{u}}r}]{Goel:19}
Rahul Goel, Shachi Paul, and Dilek Hakkani{-}T{\"{u}}r. 2019.
\newblock Hyst: {A} hybrid approach for flexible and accurate dialogue state
  tracking.
\newblock In \emph{Interspeech 2019, 20th Annual Conference of the
  International Speech Communication Association, Graz, Austria, 15-19
  September 2019}, pages 1458--1462. {ISCA}.

\bibitem[{Hashimoto et~al.(2017)Hashimoto, Xiong, Tsuruoka, and
  Socher}]{Hashimoto:17}
Kazuma Hashimoto, Caiming Xiong, Yoshimasa Tsuruoka, and Richard Socher. 2017.
\newblock A joint many-task model: Growing a neural network for multiple {NLP}
  tasks.
\newblock In \emph{Proceedings of the 2017 Conference on Empirical Methods in
  Natural Language Processing, {EMNLP} 2017, Copenhagen, Denmark, September
  9-11, 2017}, pages 1923--1933.

\bibitem[{Henderson et~al.(2014)Henderson, Thomson, and Young}]{Henderson:14}
Matthew Henderson, Blaise Thomson, and Steve~J. Young. 2014.
\newblock Word-based dialog state tracking with recurrent neural networks.
\newblock In \emph{Proceedings of the {SIGDIAL} 2014 Conference, The 15th
  Annual Meeting of the Special Interest Group on Discourse and Dialogue, 18-20
  June 2014, Philadelphia, PA, {USA}}, pages 292--299.

\bibitem[{Huang et~al.(2020)Huang, Feng, Hu, Wu, Du, and Ma}]{Huang:20}
Yi~Huang, Junlan Feng, Min Hu, Xiaoting Wu, Xiaoyu Du, and Shuo Ma. 2020.
\newblock Meta-reinforced multi-domain state generator for dialogue systems.
\newblock In \emph{Proceedings of the 58th Annual Meeting of the Association
  for Computational Linguistics, {ACL} 2020, Online, July 5-10, 2020}, pages
  7109--7118. Association for Computational Linguistics.

\bibitem[{Jang et~al.(2016)Jang, Ham, Lee, Chang, and Kim}]{Jang:16}
Youngsoo Jang, Jiyeon Ham, Byung{-}Jun Lee, Youngjae Chang, and Kee{-}Eung Kim.
  2016.
\newblock Neural dialog state tracker for large ontologies by attention
  mechanism.
\newblock In \emph{2016 {IEEE} Spoken Language Technology Workshop, {SLT} 2016,
  San Diego, CA, USA, December 13-16, 2016}, pages 531--537.

\bibitem[{Kingma and Ba(2015)}]{Kingma:15}
Diederik~P. Kingma and Jimmy Ba. 2015.
\newblock Adam: {A} method for stochastic optimization.
\newblock In \emph{3rd International Conference on Learning Representations,
  {ICLR} 2015, San Diego, CA, USA, May 7-9, 2015, Conference Track
  Proceedings}.

\bibitem[{Lee and Kim(2016)}]{Lee:16}
Byung{-}Jun Lee and Kee{-}Eung Kim. 2016.
\newblock Dialog history construction with long-short term memory for robust
  generative dialog state tracking.
\newblock \emph{D{\&}D}, 7(3):47--64.

\bibitem[{Lee et~al.(2019)Lee, Lee, and Kim}]{Lee:19}
Hwaran Lee, Jinsik Lee, and Tae{-}Yoon Kim. 2019.
\newblock {SUMBT:} slot-utterance matching for universal and scalable belief
  tracking.
\newblock In \emph{{ACL} 2019}, pages 5478--5483.

\bibitem[{Liu and Perez(2017)}]{Liu:17a}
Fei Liu and Julien Perez. 2017.
\newblock Dialog state tracking, a machine reading approach using memory
  network.
\newblock In \emph{Proceedings of the 15th Conference of the European Chapter
  of the Association for Computational Linguistics, {EACL} 2017, Valencia,
  Spain, April 3-7, 2017, Volume 1: Long Papers}, pages 305--314.

\bibitem[{Mrksic et~al.(2015)Mrksic, S{\'{e}}aghdha, Thomson, Gasic, Su,
  Vandyke, Wen, and Young}]{Mrksic:15}
Nikola Mrksic, Diarmuid~{\'{O}} S{\'{e}}aghdha, Blaise Thomson, Milica Gasic,
  Pei{-}hao Su, David Vandyke, Tsung{-}Hsien Wen, and Steve~J. Young. 2015.
\newblock Multi-domain dialog state tracking using recurrent neural networks.
\newblock In \emph{Proceedings of the 53rd Annual Meeting of the Association
  for Computational Linguistics and the 7th International Joint Conference on
  Natural Language Processing of the Asian Federation of Natural Language
  Processing, {ACL} 2015, July 26-31, 2015, Beijing, China, Volume 2: Short
  Papers}, pages 794--799.

\bibitem[{Mrksic et~al.(2017)Mrksic, S{\'{e}}aghdha, Wen, Thomson, and
  Young}]{Mrksic:17}
Nikola Mrksic, Diarmuid~{\'{O}} S{\'{e}}aghdha, Tsung{-}Hsien Wen, Blaise
  Thomson, and Steve~J. Young. 2017.
\newblock Neural belief tracker: Data-driven dialogue state tracking.
\newblock In \emph{Proceedings of the 55th Annual Meeting of the Association
  for Computational Linguistics, {ACL} 2017, Vancouver, Canada, July 30 -
  August 4, Volume 1: Long Papers}, pages 1777--1788.

\bibitem[{Nouri and Hosseini{-}Asl(2018)}]{Nouri:18}
Elnaz Nouri and Ehsan Hosseini{-}Asl. 2018.
\newblock Toward scalable neural dialogue state tracking model.
\newblock \emph{CoRR}, abs/1812.00899.

\bibitem[{Pennington et~al.(2014)Pennington, Socher, and
  Manning}]{Pennington:14}
Jeffrey Pennington, Richard Socher, and Christopher~D. Manning. 2014.
\newblock Glove: Global vectors for word representation.
\newblock In \emph{Proceedings of the 2014 Conference on Empirical Methods in
  Natural Language Processing, {EMNLP} 2014, October 25-29, 2014, Doha, Qatar,
  {A} meeting of SIGDAT, a Special Interest Group of the {ACL}}, pages
  1532--1543.

\bibitem[{Ramadan et~al.(2018)Ramadan, Budzianowski, and Gasic}]{Ramadan:18}
Osman Ramadan, Pawel Budzianowski, and Milica Gasic. 2018.
\newblock Large-scale multi-domain belief tracking with knowledge sharing.
\newblock In \emph{Proceedings of the 56th Annual Meeting of the Association
  for Computational Linguistics, {ACL} 2018, Melbourne, Australia, July 15-20,
  2018, Volume 2: Short Papers}, pages 432--437.

\bibitem[{Rastogi et~al.(2017)Rastogi, Hakkani{-}T{\"{u}}r, and
  Heck}]{Rastogi:17}
Abhinav Rastogi, Dilek Hakkani{-}T{\"{u}}r, and Larry~P. Heck. 2017.
\newblock Scalable multi-domain dialogue state tracking.
\newblock In \emph{2017 {IEEE} Automatic Speech Recognition and Understanding
  Workshop, {ASRU} 2017, Okinawa, Japan, December 16-20, 2017}, pages 561--568.

\bibitem[{Ren et~al.(2018)Ren, Xie, Chen, and Yu}]{Ren:18}
Liliang Ren, Kaige Xie, Lu~Chen, and Kai Yu. 2018.
\newblock Towards universal dialogue state tracking.
\newblock In \emph{Proceedings of the 2018 Conference on Empirical Methods in
  Natural Language Processing, Brussels, Belgium, October 31 - November 4,
  2018}, pages 2780--2786.

\bibitem[{Serban et~al.(2016)Serban, Sordoni, Bengio, Courville, and
  Pineau}]{Serban:16}
Iulian~Vlad Serban, Alessandro Sordoni, Yoshua Bengio, Aaron~C. Courville, and
  Joelle Pineau. 2016.
\newblock Building end-to-end dialogue systems using generative hierarchical
  neural network models.
\newblock In \emph{Proceedings of the Thirtieth {AAAI} Conference on Artificial
  Intelligence, February 12-17, 2016, Phoenix, Arizona, {USA}}, pages
  3776--3784. {AAAI} Press.

\bibitem[{Serban et~al.(2017)Serban, Sordoni, Lowe, Charlin, Pineau, Courville,
  and Bengio}]{Serban:17}
Iulian~Vlad Serban, Alessandro Sordoni, Ryan Lowe, Laurent Charlin, Joelle
  Pineau, Aaron~C. Courville, and Yoshua Bengio. 2017.
\newblock A hierarchical latent variable encoder-decoder model for generating
  dialogues.
\newblock In \emph{Proceedings of the Thirty-First {AAAI} Conference on
  Artificial Intelligence, February 4-9, 2017, San Francisco, California,
  {USA}}, pages 3295--3301. {AAAI} Press.

\bibitem[{Shi et~al.(2016)Shi, Ushio, Endo, Yamagami, and Horii}]{Shi:16}
Hongjie Shi, Takashi Ushio, Mitsuru Endo, Katsuyoshi Yamagami, and Noriaki
  Horii. 2016.
\newblock Convolutional neural networks for multi-topic dialog state tracking.
\newblock In \emph{Dialogues with Social Robots - Enablements, Analyses, and
  Evaluation, Seventh International Workshop on Spoken Dialogue Systems,
  {IWSDS} 2016, Saariselk{\"{a}}, Finland, January 13-16, 2016}, pages
  451--463.

\bibitem[{Thomson and Young(2010)}]{Thomson:10}
Blaise Thomson and Steve~J. Young. 2010.
\newblock Bayesian update of dialogue state: {A} {POMDP} framework for spoken
  dialogue systems.
\newblock \emph{Computer Speech {\&} Language}, 24(4):562--588.

\bibitem[{Vinyals and Le(2015)}]{Vinyals:15}
Oriol Vinyals and Quoc~V. Le. 2015.
\newblock A neural conversational model.
\newblock \emph{CoRR}, abs/1506.05869.

\bibitem[{Vodol{\'{a}}n et~al.(2017)Vodol{\'{a}}n, Kadlec, and
  Kleindienst}]{Vodol:17}
Miroslav Vodol{\'{a}}n, Rudolf Kadlec, and Jan Kleindienst. 2017.
\newblock Hybrid dialog state tracker with {ASR} features.
\newblock In \emph{Proceedings of the 15th Conference of the European Chapter
  of the Association for Computational Linguistics, {EACL} 2017, Valencia,
  Spain, April 3-7, 2017, Volume 2: Short Papers}, pages 205--210.

\bibitem[{Wang and Lemon(2013)}]{Wang:13}
Zhuoran Wang and Oliver Lemon. 2013.
\newblock A simple and generic belief tracking mechanism for the dialog state
  tracking challenge: On the believability of observed information.
\newblock In \emph{Proceedings of the {SIGDIAL} 2013 Conference, The 14th
  Annual Meeting of the Special Interest Group on Discourse and Dialogue, 22-24
  August 2013, SUPELEC, Metz, France}, pages 423--432.

\bibitem[{Williams and Zipser(1989)}]{Williams:89}
Ronald~J. Williams and David Zipser. 1989.
\newblock A learning algorithm for continually running fully recurrent neural
  networks.
\newblock \emph{Neural Computation}, 1(2):270--280.

\bibitem[{Wu et~al.(2019)Wu, Madotto, Hosseini{-}Asl, Xiong, Socher, and
  Fung}]{Wu:19}
Chien{-}Sheng Wu, Andrea Madotto, Ehsan Hosseini{-}Asl, Caiming Xiong, Richard
  Socher, and Pascale Fung. 2019.
\newblock Transferable multi-domain state generator for task-oriented dialogue
  systems.
\newblock In \emph{Proceedings of the 57th Conference of the Association for
  Computational Linguistics, {ACL} 2019, Florence, Italy, July 28- August 2,
  2019, Volume 1: Long Papers}, pages 808--819.

\bibitem[{Xu and Hu(2018)}]{Xu:18}
Puyang Xu and Qi~Hu. 2018.
\newblock An end-to-end approach for handling unknown slot values in dialogue
  state tracking.
\newblock In \emph{Proceedings of the 56th Annual Meeting of the Association
  for Computational Linguistics, {ACL} 2018, Melbourne, Australia, July 15-20,
  2018, Volume 1: Long Papers}, pages 1448--1457.

\bibitem[{Young et~al.(2010)Young, Gasic, Keizer, Mairesse, Schatzmann,
  Thomson, and Yu}]{Young:10}
Steve~J. Young, Milica Gasic, Simon Keizer, Fran{\c{c}}ois Mairesse, Jost
  Schatzmann, Blaise Thomson, and Kai Yu. 2010.
\newblock The hidden information state model: {A} practical framework for
  pomdp-based spoken dialogue management.
\newblock \emph{Computer Speech {\&} Language}, 24(2):150--174.

\bibitem[{Yu et~al.(2015)Yu, Sun, Chen, and Zhu}]{Yu:15}
Kai Yu, Kai Sun, Lu~Chen, and Su~Zhu. 2015.
\newblock Constrained markov bayesian polynomial for efficient dialogue state
  tracking.
\newblock \emph{{IEEE/ACM} Trans. Audio, Speech {\&} Language Processing},
  23(12):2177--2188.

\bibitem[{Zhang et~al.(2019)Zhang, Hashimoto, Wu, Wan, Yu, Socher, and
  Xiong}]{Zhang:19}
Jianguo Zhang, Kazuma Hashimoto, Chien{-}Sheng Wu, Yao Wan, Philip~S. Yu,
  Richard Socher, and Caiming Xiong. 2019.
\newblock Find or classify? dual strategy for slot-value predictions on
  multi-domain dialog state tracking.
\newblock \emph{CoRR}, abs/1910.03544.

\bibitem[{Zhong et~al.(2018)Zhong, Xiong, and Socher}]{Zhong:18}
Victor Zhong, Caiming Xiong, and Richard Socher. 2018.
\newblock Global-locally self-attentive dialogue state tracker.
\newblock In \emph{Proceedings of the 56th Annual Meeting of the Association
  for Computational Linguistics, {ACL} 2018, Melbourne, Australia, July 15-20,
  2018, Volume 1: Long Papers}.

\bibitem[{Zilka and Jurc{\'{\i}}cek(2015)}]{Zilka:15}
Luk{\'{a}}s Zilka and Filip Jurc{\'{\i}}cek. 2015.
\newblock Incremental lstm-based dialog state tracker.
\newblock In \emph{2015 {IEEE} Workshop on Automatic Speech Recognition and
  Understanding, {ASRU} 2015, Scottsdale, AZ, USA, December 13-17, 2015}, pages
  757--762.

\end{thebibliography}

\end{document}